\documentclass[10pt,twocolumn,letterpaper]{article}

\usepackage{iccv}
\usepackage{times}
\usepackage{epsfig}
\usepackage{graphicx}
\usepackage{amsmath}
\usepackage{amssymb}
\usepackage{booktabs}

\usepackage{multirow}
\usepackage{algorithm,algorithmicx,algpseudocode}

\usepackage{color}

\usepackage[pagebackref=true,breaklinks=true,letterpaper=true,colorlinks,bookmarks=false]{hyperref}

 \iccvfinalcopy 

\def\iccvPaperID{1897} 

\definecolor{orange}{rgb}{1,0.5,0}

\ificcvfinal\fi

\begin{document}

\title{Object Detection in Video with Spatial-temporal Context Aggregation}
\author{Hao Luo$^{\dag}$$^*$ \ \  Lichao Huang$^{\ddag}$ \ \  Han Shen$^{\ddag}$ \ \ Yuan Li$^{\ddag}$ \ \   Chang Huang$^{\ddag}$ \ \  Xinggang Wang$^{\dag}$  \\
        $^{\dag}$School of EIC, Huazhong University of Science and Technology \\
        $^{\ddag}$Horizon Robotics Inc. \\
    {\tt\small \{luohao,xgwang\}@hust.edu.cn \{lichao.huang,han.shen,yuan.li,chang.huang\}@horizon.ai}
}

\maketitle

\begin{abstract}
   Recent cutting-edge feature aggregation paradigms for video object detection rely on inferring feature correspondence. The feature correspondence estimation problem is fundamentally difficult due to poor image quality, motion blur, \etc, and the results of feature correspondence estimation are unstable. To avoid the problem, we propose a simple but effective feature aggregation framework which operates on the object proposal-level. It learns to enhance each proposal's feature via modeling semantic and spatio-temporal relationships among object proposals from both within a frame and across adjacent frames. Experiments are carried out on the ImageNet VID dataset \cite{deng2009imagenet}. Without any bells and whistles, our method obtains 80.3\% mAP on the ImageNet VID dataset, which is superior over the previous state-of-the-arts. The proposed feature aggregation mechanism improves the single frame Faster RCNN baseline by 5.8\% mAP. Besides, under the setting of no temporal post-processing, our method outperforms the previous state-of-the-art by 1.4\% mAP.
\end{abstract}

\let\thefootnote\relax\footnote{$^*$~The work was done when Hao Luo was an intern in Horizon Robotics Inc.}

\section{Introduction}

Tremendous progress has been made on object detection in static images \cite{redmon2016you, he2017mask, ren2015faster, huang2015densebox} thanks to the success of deep convolutional neural networks (CNN). However, object detection in video remains a much more challenging problem since videos often suffer from image quality degeneration such as motion blur and video defocus, which has not been addressed adequately by the state-of-the-art image-level detectors. On the other hand, videos have richer context information in both spatial and temporal domains, making it crucial to incorporate spatio-temporal context to improve detection accuracy. Feature aggregation is one important technique that has been proven effective in many video related tasks, \eg, video action recognition \cite{kar2017adascan, karpathy2014large}.


Following this direction, several previous approaches \cite{zhu2017flow, bertasius2018object, wang2018fully} apply feature aggregation to video object detection by learning a pixel-level aggregation strategy to improve the feature of a frame based on the information from its adjacent frames. In addition, feature calibration is done using motion information explicitly produced by optical flow or FlowNet \cite{dosovitskiy2015flownet}, or implicitly modeled by deformable convolution network (DCN) \cite{dai2017deformable}. 

\begin{figure}[t]
\label{fig:intro}
\centering
\footnotesize
\includegraphics[width=0.95\linewidth]{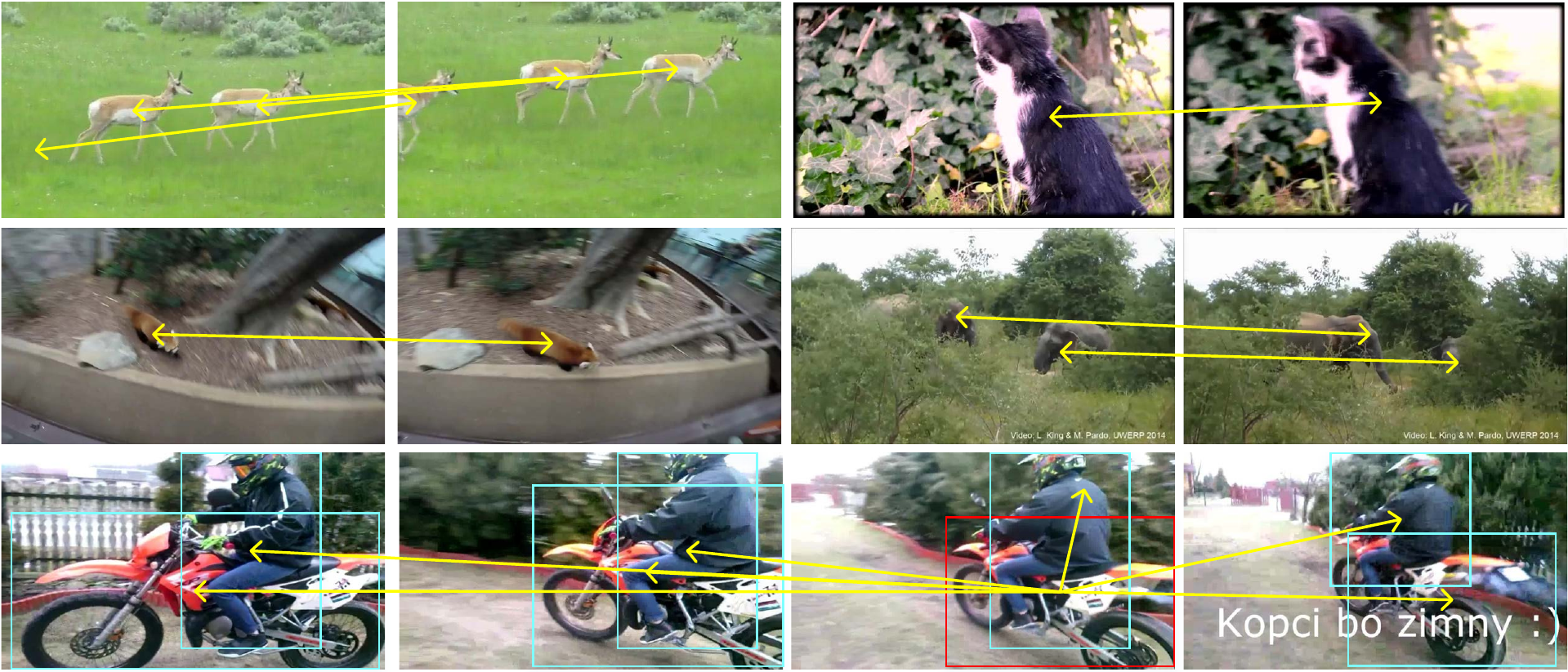}
\caption{Illustration of the differences between pixel/instance-level feature correspondence (the top and middle rows) and the proposed proposal-level spatio-temporal context (the bottom row). The former easily suffers from emergence of new objects, video defocus, motion blur, occlusion \etc. while the latter captures the dependency among proposals of intra- and inter-frames.}
\end{figure}

However, a range of problems such as motion blur, illumination variation, occlusion, drastic translation and scale change \etc can make pixel-wise motion estimation error-prone and jeopardize the above pixel-level feature aggregation methods, as illustrated in top and middle row in Fig.~\ref{fig:intro}. Facing the these problems,  the feature calibration strategy in pixel-level feature propagation methods attempt to solve these problems but brings extra problems. Motion information predicted by CNN (\eg, FlowNet and DCN) and other non-deep methods may be inaccurate, which will make the feature calibration illogical.
These ambiguous features obtained via pixel-level feature propagation will confuse the detection network. With a task-specific loss in training, these problems are alleviated to some extent but not solved fundamentally. Besides, the above feature calibration methods usually focus only on the same object instance, while overlooking contextual relations among different object instances or categories. For example, pixel-wise feature aggregation cannot capture things like a mouse is likely to sit besides a computer and a boat is rarely on a lawn, which is no doubt beneficial to detection accuracy.  From this perspective, capturing feature correspondence between frames is suboptimal for feature aggregation. 

\begin{figure*}[ht]
\centering
\footnotesize
\includegraphics[width=0.95\linewidth]{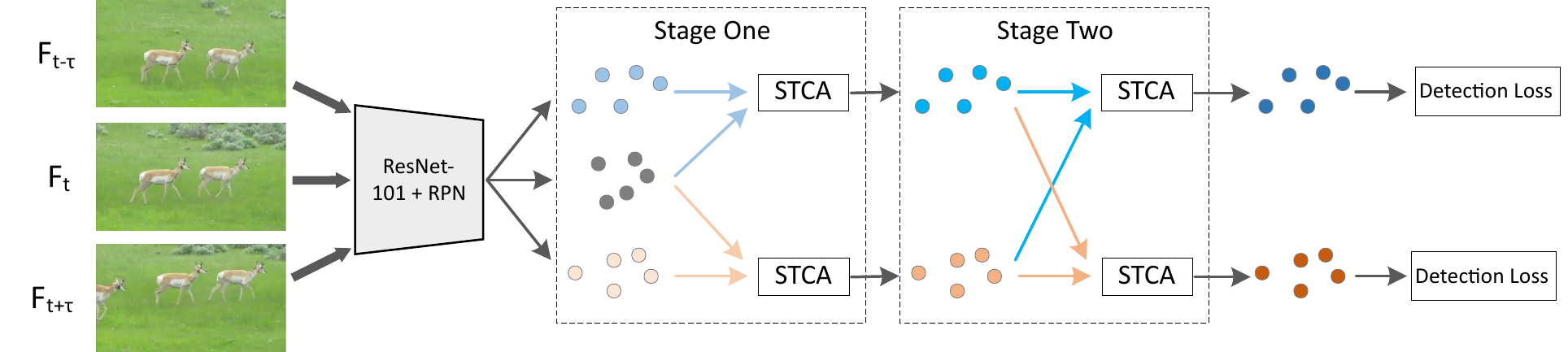}
\caption{Proposed spatial-temporal feature aggregation framework. For each input video frame, ResNet-101 is used to extract the feature map, followed by RPN which generates object proposals. Each dot represents the semantic feature of the corresponding proposal. The STCA operator is then applied in two stages to enhances each proposal's feature by using multiple related proposals' semantic features, spatial coordinates and temporal positions as input, as elaborated in Eq.~\eqref{stca}.}

\label{fig:framework}
\end{figure*}

In this paper, we focus on improving the video object detection accuracy by object-level \textit{spatial-temporal context aggregation} (STCA), as illustrated in the bottom row of Fig.~\ref{fig:intro}. A novel network is proposed to enhance the feature of each object proposal by aggregating features of other object proposals within the same fame and across neighboring frames based on spatio-temporal relations. More precisely, we utilize the region proposal network (RPN) of \cite{ren2015faster} to generate proposals for each frame separately and extract a corresponding semantic feature for each proposal via ROI-Pooling \cite{girshick2015fast}. These semantic features are input to a proposal-level feature aggregation unit, which automatically learns the dependency between proposal features and aggregates them to form new proposal representations for more accurate recognition. Within the aggregation unit, the proposal features are regarded as tokens and the self-attention mechanism \cite{vaswani2017attention} is utilized to capture the dependency among them. 
The proposed feature-based dependency modeling focuses on the visual content of image region only and the position information among proposals is ignored. Hence, we explicitly add spatial position representation to the attention for modeling the geometric information with the region proposals in one image. Besides of the spatial position information, the temporal position information among proposals across time is also captured in our method. 
Then the proposed context dependency model is both content- and position-aware, which makes full use of spatial-temporal context. Proposal-level feature aggregation is performed according to the learned dependency. To make the feature expressive enough, we exploit STCA units as in Fig.~\ref{fig:framework}. Finally, the enhanced object feature is fed as the input to subsequent object detection module \ie, the head net of Faster R-CNN. All steps are integrated in the proposed aggregation network and it can be trained in an end-to-end manner. 

Compared with existing pixel-wise feature aggregation methods \cite{zhu2017flow, bertasius2018object, wang2018fully}, our proposal-level STCA has the following merits: (1) STCA does not have any hand-crafted design, \eg, the feature wrapping process in pixel-level feature propagation; it is fully trainable; (2) STCA circumvents the challenging problem of accurate feature correspondence estimation across video frames, which makes it robust to low quality image frames. (3) STCA can capture both intra-frame contextual relations among different objects and inter-frame spatio-temporal relations, which has not been achieved in any previous video object detection method. (4) In the experiments, based on the single-frame Faster RCNN, a two-tier STCA obtains the state-of-the-art performance on ImageNet VID without any temporal post-processing.


\section{Related Work}

\vspace{.1cm}
\paragraph{Object detection in static Images.}
Deep CNN based methods dominate quite a few computer vision tasks, including object detection \cite{girshick2014rich, girshick2015fast, dai2016r, ren2015faster}. State-of-the-art image-level object detection algorithms are defined as two types: one-stage detector and two-stage detector. Two-stage detectors mainly follow the R-CNN pipeline. R-CNN~\cite{girshick2014rich} first proposes to use CNN to extract region features and then input them to a subsequent recognition network, which achieves a remarkable performance gain in comparison with traditional methods. However, it is time and space consuming. To solve these problems, Fast R-CNN~\cite{girshick2015fast} proposes to share computation for all proposals and directly conducts ROI-Pooling on CNN feature maps of the whole image. To integrate all stages and make detector's training end-to-end, RPN~\cite{ren2015faster} is proposed to replace the hand-crafted proposal generation stage. Lately, a mass of improved approaches have emerged, \eg Mask R-CNN \cite{he2017mask} and FPN \cite{lin2017feature}, which further boost the recognition accuracy. By contrast, one-stage detector is usually faster and simpler but may hinder the performance a little bit. YOLO \cite{redmon2016you} is one of the classical one-stage detectors, which divides image into grids and outputs prediction at each grid cell. Without supernumerary proposal generation and region refinement, YOLO can run in real-time. To handle with object scale variation, SSD \cite{liu2016ssd} executes detection on image's feature map of various resolution. RetinaNet \cite{lin2017focal} concentrates on the class imbalance problem. Equipped with well-designed techniques, modern one-stage detectors can obtain similar accuracy as two-stage detectors. On the whole, object detection in static images gets great progress. Besides, modeling contextual information within an image for better object is also an active topic. For example,  the relation network \cite{hu2018relation} uses self-attention \cite{vaswani2017attention} to explorer the relation between detection hypothesis for duplicate removal and obtains better better detection accuracy at the same time.  



However, video usually suffers from image quality degeneration in a real-life scenario like autonomous driving. If temporal context information is disregarded, even state-of-the-art image-level detectors behave poor on challenges particular to video. But if temporal context information in video is taken into consideration, things get easier. Here we focus on it. We take Faster R-CNN as our basic detector and extend it for video object detection.

\vspace{.1cm}
\paragraph{Object detection in video.}
Lacking large-scale annotated video datasets, video object detection benefits less from deep learning until the emergence of the ImageNet VID dataset. After that, there are increasing work exploring how to exploit temporal context for video object detection. Roughly, it falls into two categories, box score propagation and feature propagation \cite{tang2018object}. 
Box score propagation concentrates on suppressing false positive examples and boosting the confidence of true positive examples. In T-CNN \cite{kang2018t}, they first obtain region proposals of the whole video clip by an image-level detector. Then optical flow is exploited to propagate proposal across neighboring frames and tubelet is generated by tracker. Finally, box score propagation within tubelet is carried out. 
Discriminatively, in \cite{tang2018object}, short tubelet is generated firstly via CNN and then they propose a high quality linking technique to obtain better tubelet. Also, box score propagation is used within tubelet finally. Seq-NMS \cite{han2016seq} does box score propagation through dynamic programming. On the other hand, DFF \cite{zhu2017deep} utilizes optical flow to propagate feature across frames to avoid costly feature extraction for non-key frame. Optical flow is predicted by FlowNet \cite{dosovitskiy2015flownet}. FGFA \cite{zhu2017flow} propagates feature in identical way and furthermore, a pixel-wise feature aggregation is learned. Sharing the same spirit, STSN \cite{bertasius2018object} is the same thing but substitutes deformable convolution for optical flow's prediction. Based on FGFA, MANet~\cite{wang2018fully} adds extra object-level calibration to do feature propagation but additional annotation of instance id is of need. They take optical flow and proposal's coordinates as input and output relative movements of identical instance. Eventually, only prediction of identical instance across consecutive frames is directly averaged. All these methods utilize temporal information to regularize detection results via temporal post-processing techniques. 

Our method belongs to the latter, feature propagation in some extent. MANet \cite{wang2018fully} is partly similar to our work where proposal-level aggregation is adopted. Yet completely different from it, we propose to enhance proposal-level feature via a learnable spatial-temporal aggregation unit, which takes feature of all contexts region intra- and inter-frames into account and regularize the detection results in the training. And moreover, no additional data annotations, \eg instance id or optical flow, are required.

\vspace{.1cm}
\paragraph{Attention for sequence modeling.}
In the past few years, attention mechanisms have witnessed many distinct advances in sequence modeling and now it even becomes one of the standard component in sequence modeling. Self-attention (means Scaled Dot-Product Attention \cite{vaswani2017attention}) reveals notable advantages, including longer range dependency, more parallelizable and lower computational complexity, in comparison with the counterpart, RNN. In consideration of self-attention's powerful capacity for context embedding and the similarity between general sequence and video, we modulate self-attention \cite{vaswani2017attention} to model dependency among context regions within video clip for video object detection task. Self-attention has also been applied for video classification as the non-local network in \cite{wang2018non}. 


\section{Method}

\subsection{Architecture Overview}

We aim to maximize the cooperation of spatial and temporal context for video object detection task. The overall architecture is shown in Fig.~\ref{fig:framework}. Faster R-CNN \cite{ren2015faster} is a classical static image detector comprised of a feature extractor, region proposal network (RPN) and region-based detector. Our approach is established on it with shared feature extractor and RPN. In more detail, the proposed framework takes a set of three neighboring frames $\{F_{t-\tau}, F_{t}, F_{t+\tau}\}$ within video as input. Each frame is firstly forwarded through the feature extractor (\eg, ResNet-101 \cite{he2016deep}) to obtain convolutional feature maps. Then the followed RPN is applied to generate $N$ proposals for each frame separately. Based on these proposals and feature maps, we can obtain semantic feature for each proposal of each frame by ROI-Pooling. For object detection in image, classification and regression for each proposal are conducted solely by two followed fully connected layers in Faster R-CNN. Nevertheless, for object detection in video, ignoring temporal context is not rational as commented earlier. Accordingly, we take the proposals from all the input frames into consideration and modulate this architecture with a spatial-temporal context aggregation (STCA) unit, which models the relationship explicitly between proposals from intra- and inter-frame. This unit takes semantic feature, spatial position and temporal position as input and output enhanced feature for each proposal.
The enhanced feature is context-aware in both spatial and temporal domains, which is of great importance for classification and regression. As a result, recognition accuracy will be improved. In the next sections, we will give a detailed description for the spatial-temporal context-aware unit.
\subsection{Self-attention for Semantic Dependency Modeling}

The effectiveness of self-attention in capturing short- and long-range dependency via calculating attention weight between each pair of tokens in NLP task has been proven in \cite{vaswani2017attention}. In computer vision, \cite{wang2018non} and \cite{hu2018relation} have successfully applied it to video classification and object detection in static image respectively. We further extend it to modeling the intra- and inter-frame semantic dependency among proposals for object detection in video.

Given two key frames $F_{k} = \{ F_{t-\tau}$, $F_{t+\tau} \}$ and one supporting frame $F_s = \{F_{t}\}$, the corresponding semantic features of $i$-th proposal are denoted as $\mathbf{f}^{i}_{k}$ and $\mathbf{f}^{i}_s \in \mathbb{R}^{1 \times d_v}$ respectively. In the initial spatial-temporal aggregation unit, we divide them into two groups, $G_{t, t-\tau}=\{ \mathbf{f}^{i}_{t-\tau}, \mathbf{f}^{i}_{t} \}_{i=1}^{N}$ and $G_{t, t+\tau}=\{ \mathbf{f}^{i}_{t}, \mathbf{f}^{i}_{t+\tau} \}_{i=1}^{N}$, where N is the total number of proposals per frame. Below we explain how attention is calculated for $G_{t, t-\tau}$ ($G_{t, t+\tau}$ is similar). The content-based attention weight $e^c_{ij}$ between $i$-th target proposal in key frame $F_{t-\tau}$ and $j$-th candidate proposal in $G_{t, t-\tau}$ is calculated as follows:

\begin{equation}
\label{content-based_relation}
e^c_{ij} = \frac{(\mathbf{f}^i_{t-\tau}\mathbf{W}^Q)(\mathbf{f}^j_{g}\mathbf{W}^K)^T}{\sqrt{d_v}}.
\end{equation}

To enhance the power of expression, semantic features $\mathbf{f}^i_{t-\tau}$ and $\mathbf{f}^j_{g}$ are linearly transformed via parameter matrices $\mathbf{W}^Q$, $\mathbf{W}^K \in \mathbb{R}^{d_v \times d_v}$ severally. A softmax function is applied to normalize $e^c_{ij}$ across all candidate proposals. The normalized attention weight is as follows.
\begin{equation}
\label{softmax}
w_{ij} = \frac{\exp(e_{ij})}{\sum_{m=1}^{2N}{\exp(e_{im})}}.
\end{equation}
A residual connection is adopted and eventually the enhanced feature $\mathbf{f}^i_{t-\tau, en}$ for each target proposal is as
\begin{equation}
\label{aggregation}
\mathbf{f}^i_{t-\tau, en} = \mathbf{f}^i_{t-\tau} + \sum_{j=1}^{2N}{w_{ij}\mathbf{f}_{g}^j}.
\end{equation}

In this aggregation unit, target proposals only comes from key frame and candidate proposals are from both key and supporting frames. In the same way, the enhanced feature $\mathbf{f}^i_{t+\tau, en}$ for proposals in frame $F_{t+\tau}$ can be generated as well. In the next aggregation unit, we mix all the proposals from key frames up and take their enhanced features $G_{t-\tau, t+\tau} = \{f_{t-\tau, en}, f_{t+\tau, en}\}$ as input. At the moment, every proposal will serve as a candidate proposal for each other and all parameter matrices are individual. Semantic features of proposals from key frames are updated twice and these refined features are utilized to do regression and classification.

\subsection{Spatial Position Representation}

In \cite{hu2018relation}, spatial position plays a key role in modeling spatial dependency between objects for image recognition. Video has richer spatial context by contrast with image, and so it should be of great importance as well for video object detection. Consequently, we add spatial constraints to the attention in a similar way as \cite{hu2018relation}. Concretely, given a target proposal $p^i = (x_i, y_i, w_i, h_i)$ and candidate proposal $p^j = (x_j, y_j, w_j, h_j)$, the spatial position is first transformed into $\mathbf{r} = \langle \log(\frac{|x_i - x_j|}{w_j}), \log(\frac{|y_i - y_j|}{h_j}), \log(\frac{w_i}{w_j}), \log(\frac{h_i}{h_j}) \rangle \in \mathbb{R}^4$ to make the representation invariant to scale and translation. Then sine and cosine functions of varying wavelengths is applied to each element $r \in \mathbf{r}$, as
\begin{equation}
    \label{sine_cosine}
    \begin{split}
		\boldsymbol\phi(r, 2z) &= \sin(\frac{r}{1000^{2z/d_{\phi}}}) \\
		\boldsymbol\phi(r, 2z+1) &= \cos(\frac{r}{1000^{2z/d_{\phi}}}).
	\end{split}
\end{equation}
where $z$ ranges from $0$ to $d_{\phi}/2-1$ and so each embedding vector $\boldsymbol\phi_r$ is of dimension $d_\phi$. The embedding vector $\boldsymbol\phi_\mathbf{r} \in \mathbb{R}^{1 \times 4d_{\phi}}$ of $\mathbf{r}$ is then linearly transformed by parameter matrix $W^S \in \mathbb{R}^{4d_{\phi}\times1}$. Eventually, the spatial position representation between target proposal $p^i$ and candidate proposal $p^j$ is as,
\begin{equation}
\label{eijs}
    e^s_{ij} = \boldsymbol\phi_\mathbf{r}\mathbf{W}^S.
\end{equation}

\subsection{Temporal Position Representation}

Although the spatial position representation brings the proposed network spatial constraints, it may also confuse the network without regard to temporal position. Assuming that there are two proposals with identical spatial position but from two disparate video frames, the spatial position representation between them and any other target proposals would be identical. In other words, these two proposals are treated as the same one. Hence, we incorporate temporal constraints in the attention. In video, frame number is usually not fixed, which is not suitable for modeling temporal position. Thus, we use sine and cosine functions(as in Eq.\eqref{sine_cosine}) to encode the relative temporal distance $\tau$ of frame id between two proposals $p_{t-\tau}^{i}$ and $p_{t}^{j}$. The encoded vector $\boldsymbol{\phi_\tau} \in \mathbb{R}^{1 \times d_v}$ and also it is linearly transformed by parameter matrices $\mathbf{W}^T \in \mathbb{R}^{d_v \times d_v}$. Then the temporal position representation is as,
\begin{equation}
\label{eijt}
e^t_{ij} = \frac{(\mathbf{f}^i_{t-\tau}\mathbf{W}^Q)(\boldsymbol{\phi}_\tau\mathbf{W^T})^T}{\sqrt{d_v}}.
\end{equation}

\subsection{Spatial-temporal Context Aggregation}
As introduced earlier, our STCA is responsible for the dependency modeling and feature aggregation among proposals. We unite its mathematical description here. For clarity, we denote the semantic feature and spatial-temporal position of a target proposal and a set of $M$ corresponding candidate proposals as $P^i=(\mathbf{f}^i, p^i, t^i)$ and $\mathcal{G}^i = \{{(\mathbf{f^j}, p^j, t^j)}\}_{j=1}^{M}$, respectively. The superscript $i$ represents the $i$-th proposal in key frame.

Concretely,  we can calculated  $e^c_{ij}$  following Eq.~\eqref{content-based_relation}, $e^s_{ij}$ following Eq.~\eqref{eijs} and $e^t_{ij}$ following Eq.~\eqref{eijt}. Then, the attention weight or dependency $w_{ij}$ within STCA between $p^i$ and $p^j$ is,
\begin{equation}
    \begin{split}
        e_{ij} &= e^c_{ij} + \log(e^s_{ij}) + e^t_{ij},\\
        w_{ij} &= \frac{\exp(e_{ij})}{\sum_{m=1}^{M}{\exp(e_{im})}}.
    \end{split}
\end{equation}
in which, a $\log$ function is exploited to balance the spatial position representation and other items following \cite{hu2018relation}. 
After obtaining the attention weight $w_{ij}$ between $P^i$ and each of the proposal in $\mathcal{G}^i$ , the enhanced feature of the target proposal, denoted as $\mathbf{f}^i_{en}$, can be calculated following the formulation in Eq.~\eqref{aggregation}. At last, the proposed STCA operation can be concluded as follows,
\begin{equation}
    \label{stca}
    STCA(P^i, \mathcal{G}^i) = \mathbf{f}^i_{en}.
\end{equation}



\subsection{Training \& Inference}
\label{training_and_inference}
\vspace{.1cm}
\textbf{Training.}
During training, we take Faster R-CNN\cite{ren2015faster} as our basic detector. We observe a similar detection accuracy (74.51\% mAP \vs 74.50\% mAP) between Faster R-CNN and R-FCN~\cite{dai2016r}\footnote{We use the pre-trained model released on github.}. Stride of convolutional layers in \emph{Res5} are set to 1, which changes the effective feature stride from 32 to 16. Dilated convolutions are exploited in \emph{Res5} to maintain the receptive field. A convolutional layer with 256 channels is added on the \emph{Res5} feature maps to reduce dimension. The proposed network is directly trained on the mixture of DET and VID data, see Section \ref{sec:experiment} for details. To get a triplet of images for training, one key frame is first sampled and then the other key frame and supporting frame are randomly sampled near the first key frame in range of -9 $\sim$ 9. Because DET only has images, we cannot sample sequential images strictly. In such case, we copy the image repeatedly and then take them as sequential images. Dimension $d_v$ of feature is 1024 and $d_\phi$ is 16.

\begin{algorithm}[!t]
  \small
  \caption{Inference algorithm of STCA network for video object detection }
  \label{alg:inference-framework}
  \begin{algorithmic}[1]
  \Require Sequence of video frames $\{F_k\}_{k=-2K}^{2K}$, inference window size $T = 2K + 1$.
  \Ensure Bounding boxes $B$ of frame $F_0$.
  \State{$B\gets \{\}$}
  \For{$k\gets -2K$ to $2K$} \Comment{stage 1: feature buffer}
    \State{Generate proposals $p_k$ for frame $F_k$ with RPN.}
    \State{Extract features $f_k$ for proposals $p_k$.}
    \State{Assign temporal position $t_k$ for proposals $p_k$.}
  \EndFor
  \For{$k \gets -K$ to $K$} \Comment{stage 2: feature buffer}
    \State{Prepare target proposal $P_k=(f_k, p_k, t_k)$.}
    \State{Prepare candidate proposals $\mathcal{G}_k=\{(f_i, p_i, t_i)\}_{i=k-K}^{k+K}$.}
    \State{$f_{k, en} \gets STCA(P_k, \mathcal{G}_k)$.}
  \EndFor
  \For{$k \gets -K$ to $K$} \Comment{update feature}
  \State{$f_k \gets f_{k, en}$.}
  \EndFor
  \State{$f_0 \gets STCA(P_0, \mathcal{G}_0)$.} \Comment{stage 3: detection}
  \State{Do detection with $f_0$.}
  \State{Add detection results to $B$}
  \end{algorithmic}
\end{algorithm}

\vspace{.1cm}
\textbf{Inference.}
Being consistent with training, we take proposals of key frame and adjacent supporting frames (resulting in $T$ frames in total) as candidate proposals to assist target proposals in key frame during inference. More details of the inference phase are summarized in Algorithm~\ref{alg:inference-framework}. At present, the temporal receptive field for key frame $F_0$ is $2T - 1$. It can be implemented easily, and for efficiency, we decouple it into three submodules and adopt two feature buffers to store the shared features for the first two stages, separately. For frames around the video boundary, we pad the buffer with boundary frame for convenience. Unless otherwise specified, $T$ is set to 31 by default. The proposals number $N$ of each frame is set to 300. We will investigate how performance varies with hyperparameters $T$ and $N$ in the ablation studies. NMS threshold for RPN is 0.7 and for the final detection results refinement, it is 0.5.

\section{Experiments}
\subsection{Experiment Setup}
\label{sec:experiment}

\vspace{.1cm}
\textbf{Dataset.}
Following most of previous state-of-the-art methods, we conduct all experiments on the ImageNet VID dataset \cite{deng2009imagenet} (VID). VID is a large-scale video dataset, which contains 5354 video in total. These videos are split into training, validation, and testing three subsets, which include 3862, 555, and 937 videos, respectively. Each video has about 300 frames in average and the annotations of bounding box and category for each frame are provided. There are 30 object classes in ImageNe VID and all these classes are included in the 200 classes in ImgeNet DET dataset (DET). Following the protocols in \cite{kang2018t, zhu2017deep}, images in DET overlapped with classes in VID are utilized as training data, which is a common practice. Classical detection evaluation metric Mean Average Precision (mAP) is used and all results are reported on validation set in VID because annotation data of test set is not public.

\vspace{.1cm}
\textbf{Implementation details.}
Parameters of the backbone network (ResNet-101) are initialized by a pre-trained model on ImageNet and all new layers are randomly initialized by drawing weights from a zero-mean Gaussian distribution with standard deviation 0.01. For anchors in RPN, we use 3 scales with box areas of $128^2$, $256^2$, and $512^2$ pixels, and 3 aspect ratios of 1:1, 1:2, and 2:1, resulting in 9 anchors for each spatial location. Images are resized to shorter side 600 pixels for both training and inference. We adopt SGD optimizer with momentum 0.9 and 120K iterations are trained. Learning rate is 1e-3 for first 80K iterations and 1e-4 for last 40K iterations. Weight decay is 5e-4 and batch size is set to 4, with each GPU holds one mini-batch. OHEM \cite{shrivastava2016training} with 128 rois is used. All our implementations are based on MXNet \cite{chen2015mxnet} deep learning framework.
\subsection{Ablation Study}

\begin{table}[htp]
\centering
\resizebox{\linewidth}{!}{
\begin{tabular}{c|l|c|c|l}
   \toprule
   Methods & Semantic dep. & Spatial pos. & Temporal pos. & mAP (\%) \\
   \midrule
   (a) & $(T=1)$ & & & 74.5 \\
   (b) & $(T=1)$ \ \ $\checkmark$  &&& $77.4_{\uparrow 2.9}$ \\
   (c) & $(T=31)$ $\checkmark$ & & & $79.3_{\uparrow 4.8}$ \\
   (d) & $(T=31)$ $\checkmark$ & $\checkmark$&& $79.8_{\uparrow 5.3}$ \\
   (e) & $(T=31)$ $\checkmark$ &  $\checkmark$&  $\checkmark$ & $80.3_{\uparrow 5.8}$ \\
   \bottomrule
\end{tabular}
}
\vspace{1mm}
\caption{Accuracy (mAP in $\%$) for baseline and various model variations by checking semantic dependency, spatial position and temporal position. (a) is the single-frame baseline. The inference window size $T$ of variant (b) is 1 and for variants (c)-(e), default inference window size $T$ is 31. Relative gains over the single-frame baseline (a) are listed in the subscript.}
\label{table:ablation}
\end{table}

\vspace{.1cm}
\paragraph{Model variation.}
As shown in Table~\ref{table:ablation}, to inspect the effectiveness of intra-frame and inter-frame context aggregation, as well as the proposed feature components, including semantic dependency, spatial position and temporal position, we measure the distinction of performance (mAP) on VID validation set for the  variants of our method. The detailed discussions are listed as follows.
\begin{itemize}
    \item[(a)] This is the basic Faster R-CNN single-frame baseline, which treats all video frames as static images. It is a strong baseline which obtains 74.5\% mAP using the ResNet-101 backbone. To verify the effectiveness of each component, we do not exploit techniques e.g. temporal post-processing to improve the recognition accuracy.
    \item[(b)] This setting is the basic framework of proposed method in the absence of spatial and temporal position representation. The training process is introduced as described in Section~\ref{training_and_inference}. As for the testing phase, inference window size $T$ is set to 1, namely one frame is taken as input only, which is the same as single-frame baseline (a). The corresponding mAP of method (b) is 77.4\%, which surpasses the single-frame baseline (a) by a large margin. The notable improvement (about 3\%) shows the validity of the proposed basic framework(intra-frame context aggregation with semantic feature only) for video object detection.
    \item[(c)] This method shares the same settings with method (b) but inference window size is $T = 31$. Compared with method (b), there are 1.9\% improvement in terms of mAP. This significant improvement shows that more candidate proposals from adjacent frames boost the performance. It also implies that inter-frame context could significantly improve the performance. We will explore this later in subsequent subsection.
    \item[(d)] It is established on the basic framework (c) and equipped with the spatial position representation. In despite of the excellent performance achieved in method (c), method (d) can further improve the result to 79.8\%.
    \item[(e)] This is the ultimate framework of our method. All contexts including semantic feature and spatial-temporal position are involved. It achieves an additional increase of 0.5\% compared with method (d). In the case of single-frame baseline, there are about 6\% increase, which indicates the effectiveness of our method.
\end{itemize}

\begin{figure}[htp]
\centering
\includegraphics[width=.95\linewidth]{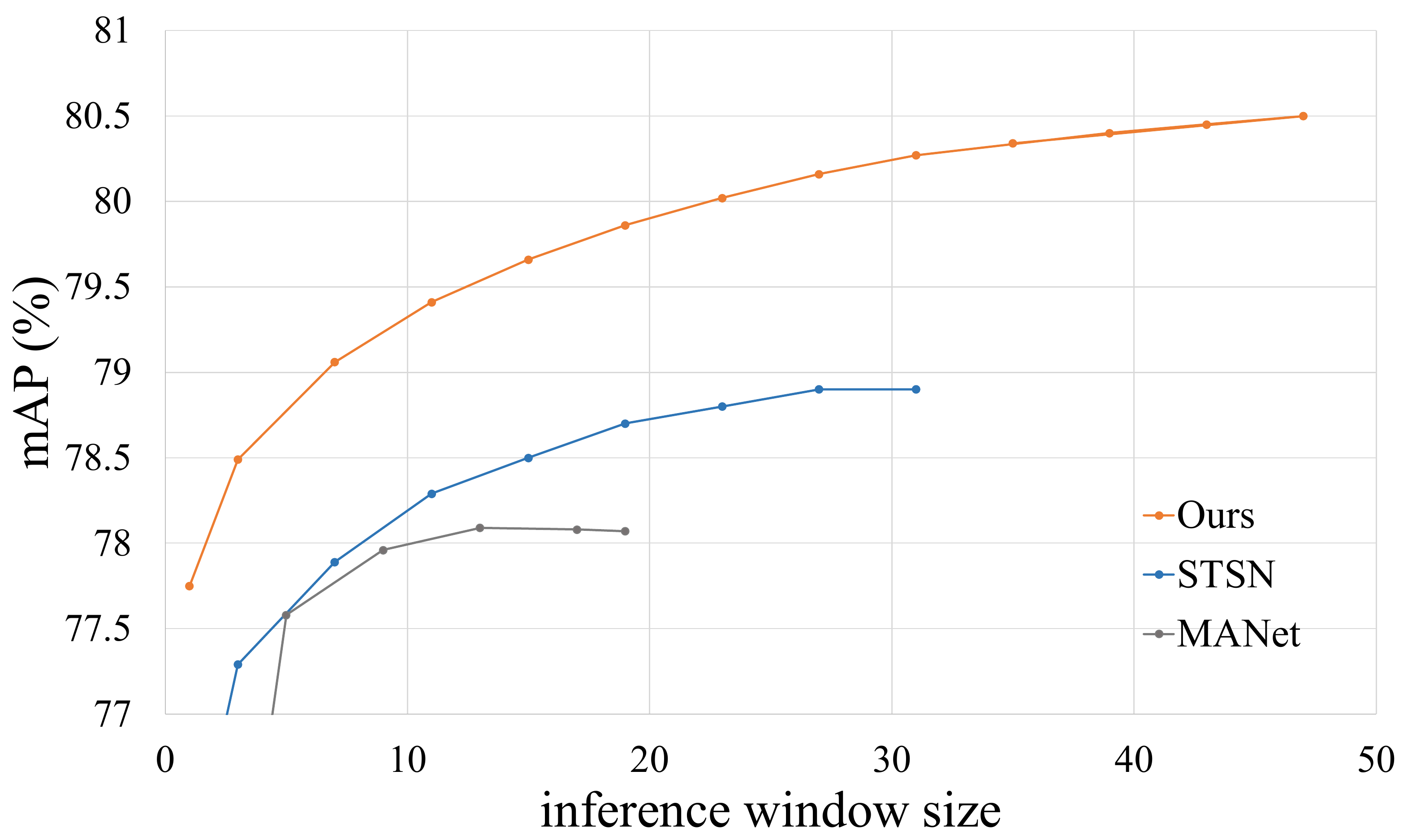}
\caption{Detection results in mAP \vs inference window size for state-of-the-art pixel-wise aggregation methods \cite{bertasius2018object, wang2018fully} and ours. We use the results reported in its corresponding papers.
}
\label{fig:ablation_inference}
\end{figure}

\begin{figure*}[!t]
\centering
\includegraphics[width=0.99\linewidth]{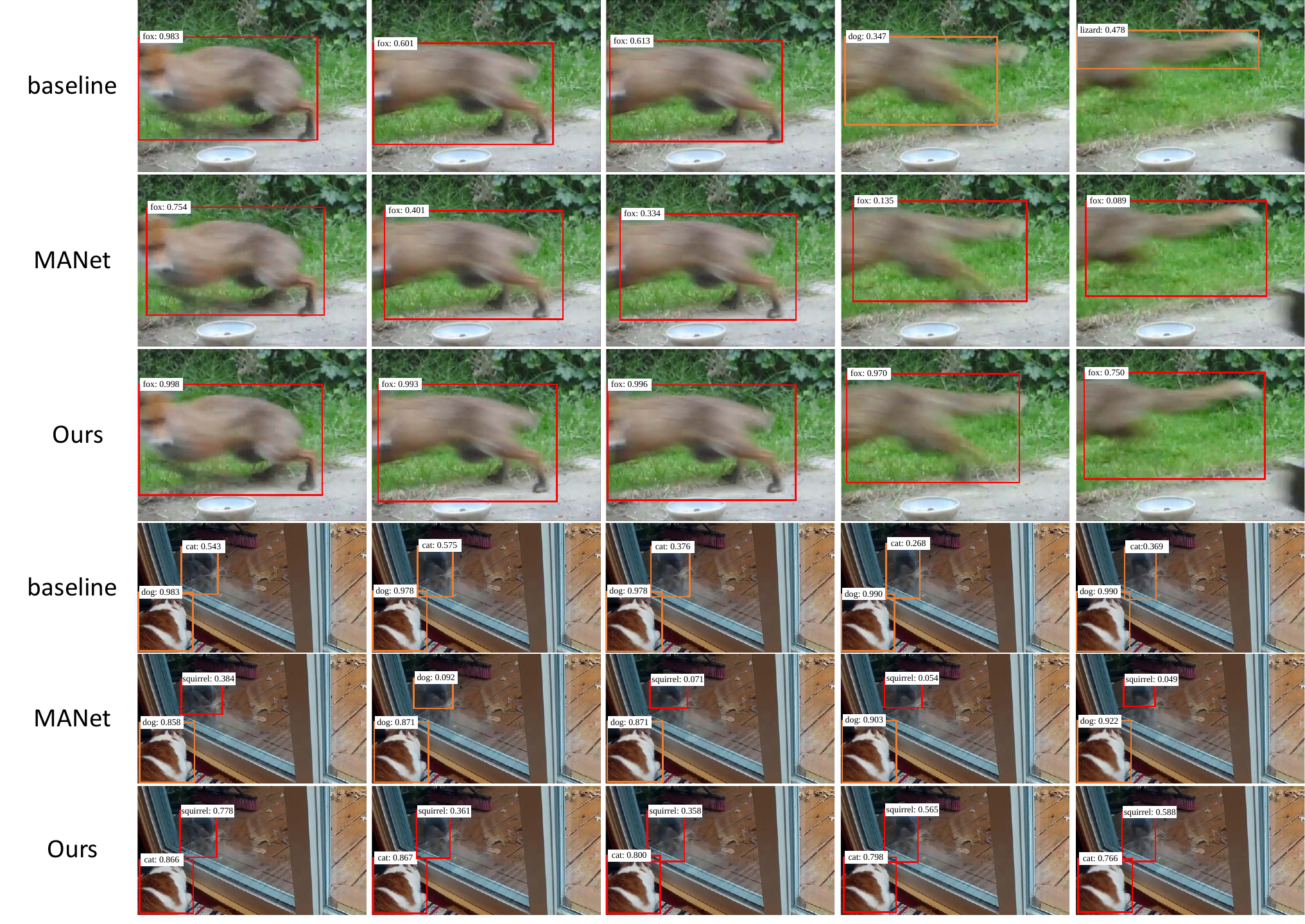}
\caption{Example detection results of single-frame baseline, MANet \cite{wang2018fully} and our method. Each row shows the results of sampled sequential video clip (only the top-scoring box around object is shown). Red and brown boxes represent correct and incorrect detection, respectively. Our method outperforms single-frame baseline and MANet when there are motion blurs and video defocus. Besides, referring to the detection confidences in the figure, superior spatial-temporal consistency can be observed in our method as well.
}
\label{fig:vis_det_res}
\end{figure*}

\begin{figure}[!t]
\centering
\footnotesize
\includegraphics[width=0.95\linewidth]{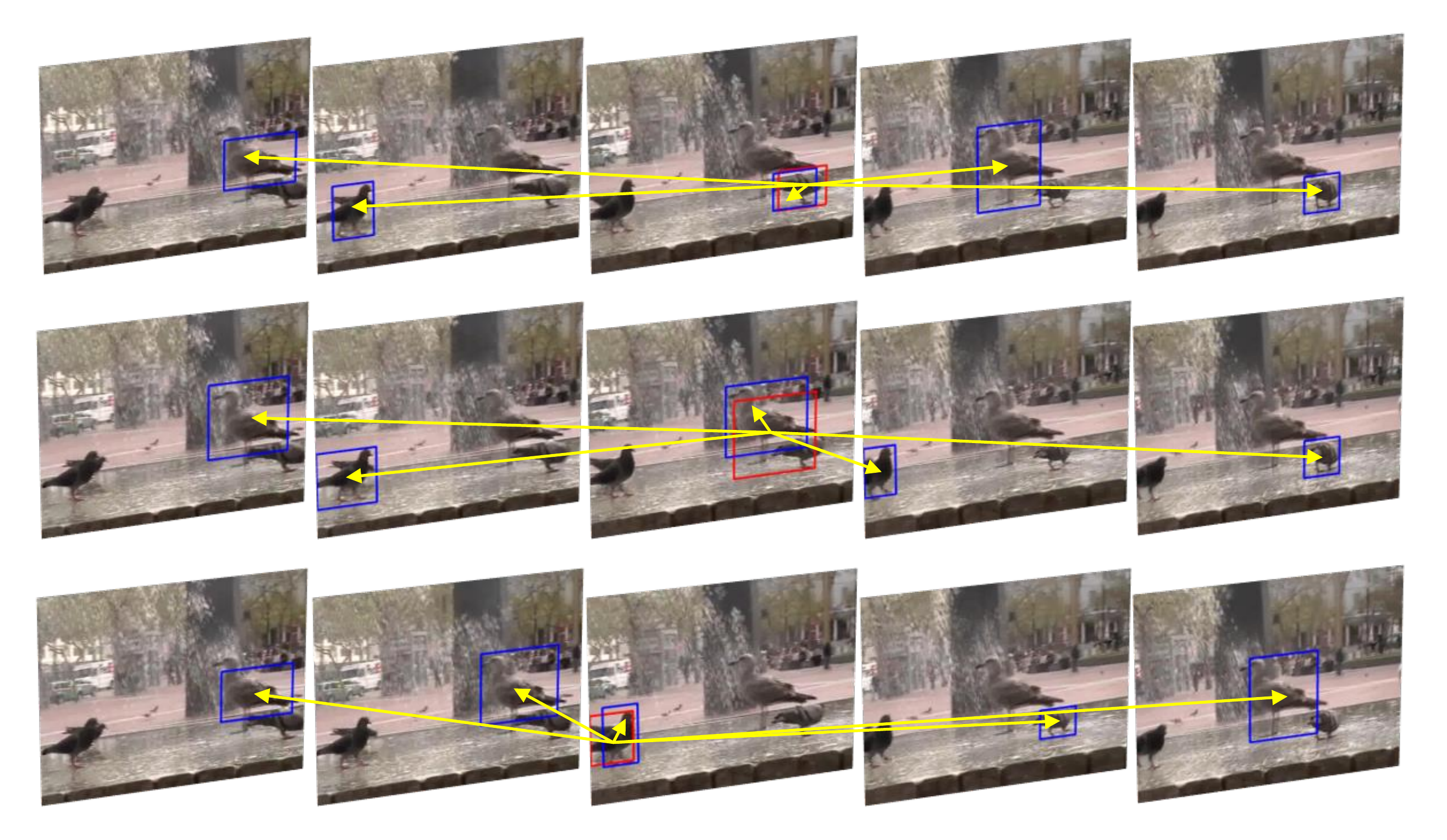}
\caption{Examples of the dependency (in Eq.~\eqref{stca}) between the target proposals and the candidate proposals from adjacent frames in the last aggregation unit. Red and blue boxes are target proposals in a key frame and candidate proposals in its neighboring frames, respectively. Each row shows one of the target proposals and its corresponding candidate proposals with highest dependency in both key frame and neighboring frames.
}
\label{fig:dependency}
\end{figure}

{
\vspace{.1cm}
\paragraph{Inference window size.} Inference window size is a key hyper-parameter and we investigate its relationship to detection results in Fig.~\ref{fig:ablation_inference}. In addition, Fig.~\ref{fig:ablation_inference} also shows the results of pixel-wise aggregation methods \cite{bertasius2018object, wang2018fully} for comparison. First, we can clearly see that the mAP of our method keeps rising with increasing the inference window size, which can be observed in other methods as well. Then MANet~\cite{wang2018fully} and STSN~\cite{bertasius2018object} stop growing at $T = $ 13 and 27, respectively. However, our method does not show obvious signs of stopping until $T = $ 40  and it reaches 80.5\% mAP when $T$ = 47. Besides, our method has the best performance all the time. It indicates that our method works well in capturing both short- and long-term dependency among proposals in spatial-temporal domain. Unless otherwise stated, the default value for $T$ is 31 and the corresponding mAP is 80.3\%.
}

{
\vspace{.1cm}
\paragraph{Number of proposals.}
We investigate how the number of proposals ($N$) for each frame affects the performance of our proposed method. The mAP is 80.3\% for $N = $ 300.  When $N$ is 128, the mAP is 79.7\%, which is slightly reduced by 0.6\% mAP. With more proposals, there are more contexts that can be mined. Hence, the number of proposals matters in our framework.
}
\begin{table}[htp]
\centering
\resizebox{\linewidth}{!}{
\begin{tabular}{c|c c c c c c c}
   \toprule
   $T$ &0& 1 & 7 & 13 & 19 & 25 & 31\\
    \midrule
    $N$ = 128 &93.6&105.7&117.3&128.5&138.8&150.4&161.0\\
    \midrule
    $N$ = 300 &96.0&113.7&153.7&195.9&238.0&279.5&322.2\\
   \bottomrule
\end{tabular}
}
\vspace{1mm}
\caption{Inference window size \vs runtime (in $ms$) for our STCA method. All the data loading and post-processing time are involved. 0 denotes the basic image detector, namely Faster R-CNN.  The runtime's evaluation is carried out on a single workstation with Titan X Maxwell GPU.}
\label{table:run_time}
\end{table}

{
\vspace{.1cm}
\paragraph{Computational overhead.}
Compared with the single-frame baseline, the additional computational overhead of the proposed method comes from the STCA unit and is mainly affected by the inference window size and proposal number. For detailed analysis, we evaluate the runtime of the proposed method with various settings as in Table~\ref{table:run_time}. On average, the newly added time cost of the STCA units is about 7.3 $ms/frame$. As for previous state-of-the-art methods, e.g, \cite{wang2018fully} and STMN \cite{xiao2018video}, the additional computation cost are 9.6 $ms/frame$ and 28 $ms/frame$, respectively. Also note that the runtime of STMN \cite{xiao2018video} and our STCA is evaluated on Titan X Maxwell GPU and for MANet \cite{wang2018fully}, more powerful Titan X Pascal GPU is utilized. Thus, our STCA is superior to MANet \cite{wang2018fully} and STMN \cite{xiao2018video} in speed as well. In addition, decreasing $N$ from 300 to 128 can greatly reduce the time cost and meanwhile the mAP decline (0.6\%) is acceptable. Thus, all the hyper-parameters of $N$ and $T$ can be adjusted within reason to achieve a good trade-off.
}

\begin{table}[htp]
\centering
\resizebox{\linewidth}{!}{
\begin{tabular}{c|cc|c}
   \toprule
   Methods & Temp. Post-Proc. & Optic Flow / ID & mAP (\%) \\
   \midrule
   \midrule
   \multirow{2}{*}{D\&T \cite{feichtenhofer2017detect}}
    && \checkmark&75.8 \\
    & \checkmark &\checkmark&79.8\\
   \midrule
   \multirow{2}{*}{ST-Lattice \cite{chen2018optimizing}}
   &&&-\\
   &\checkmark & & 79.6\\
   \midrule
   \midrule
   \multirow{2}{*}{FGFA \cite{zhu2017flow}}
    && \checkmark & 76.3\\
    & \checkmark & \checkmark & 78.4\\
   \midrule
   \multirow{2}{*}{MANet \cite{wang2018fully}}
   && \checkmark & 78.1\\
   & \checkmark & \checkmark & 80.3 \\
   \midrule
   \multirow{2}{*}{STSN \cite{bertasius2018object}}
    &&& 78.9\\
    & \checkmark && 80.4\\
   \midrule
   \multirow{2}{*}{STMN \cite{xiao2018video}}
    && & -\\
    & \checkmark && 80.5\\
   \midrule
   \midrule
   \multirow{2}{*}{STCA (Ours)}
    & & & 80.3\\
    & \checkmark && 80.6\\
   \bottomrule
\end{tabular}
}
\vspace{1mm}
\caption{We compare our proposed method with different state-of-the-are methods on the VID validation set. Detection results (mAP in $\%$) w/ and w/o temporal post-processing are reported. Temporal post-processing includes box-level techniques, \eg Seq-NMS and tubelet resocre. For STMN \cite{xiao2018video} and ST-Lattice \cite{chen2018optimizing}, only the result w/ temporal post-processing is reported in their paper. The temporal post-processing adopted for our method is Seq-NMS. ID is short for instance id.}
\label{table:detection_results}
\end{table}

\subsection{Comparison with State-of-the-art Methods}

We compare our method with the state-of-the-art pixel-wise feature aggregation methods \cite{zhu2017flow, bertasius2018object, wang2018fully, xiao2018video} and other relevant methods \cite{feichtenhofer2017detect, chen2018optimizing} for video object detection in Table~\ref{table:detection_results}. 
For illustration purposes, we make analysis on these results in terms of temporal post-processing and extra data annotation. 

\textit{Temporal post-processing.}
As is known to all, objects in videos exhibit spatial and temporal consistency. More specifically, objects in neighboring video frames would not change dramatically in both spatial and temporal domains. Hence, the detection score of box within the same tubelet should be smooth. To handle this issue, temporal post-processing techniques (\eg, Seq-NMS~\cite{han2016seq}) are proposed. Without temporal post-processing technique, STSN \cite{bertasius2018object} achieves the most excellent performance of 78.9\% mAP among all previous state-of-the-art methods. However, our method can obtain 80.3\% mAP, which is about 1.4\% higher than it. 
When temporal post-processing is applied, gain ranges from 1.5\% to 4\% for the previous state of the arts. D\&T \cite{feichtenhofer2017detect} and ST-Lattice \cite{chen2018optimizing} adopt well-designed tubelet rescore technique and others use Seq-NMS. For our case, Seq-NMS pushes our method to the state-of-the-art but the performance improvement is only 0.3\% mAP. And in our experiments, we observe the gap continue to decrease with the increase of inference window size. In some degree, this demonstrates our method can learn the temporal consistency during training and inference, which is consistent with our design, see Fig.~\ref{fig:vis_det_res} for detailed example detection results. Actually, in some realistic situations, \eg, crowded scenes, the heuristic Seq-NMS or other hand-crafted techniques may be out of action. Our method makes full use of spatial-temporal context and is trainable, which may be more promising and robust.

\textit{Extra data annotation.}
Data annotation is usually of high cost, especially for subtle annotation like optical flow. D\&T \cite{feichtenhofer2017detect} can obtains 79.8\% mAP with instance id used only. With flow annotation, FGFA~\cite{zhu2017flow} merely obtains 78.4\% mAP, which is 1.4\% lower than D\&T \cite{feichtenhofer2017detect}. MANet~\cite{wang2018fully} achieves 80.3\% mAP with both instance id and flow annotation. Without the need for any additional annotated information, our proposed method can obtain 80.6\% mAP and exceeds the best of them, which shows the superiority of our method. Also STSN~\cite{bertasius2018object} and STMN~\cite{xiao2018video} can achieve 80.4\% and 80.5\% mAP respectively under the same condition. Compared with them, our method is better as well, despite only a marginal gain. Note that the significant gain of our method is not from temporal post-processing but from the proposed framework.

Visualization of examples of the dependency among proposals from separately video frames is in Fig.~\ref{fig:dependency}. It can be observed that our STCA learns to find context regions which boost the feature quality across spatial-temporal domain in spite of the spatial or temporal distance.

\section{Discussions}

\textit{Why Seq-NMS brings less gain?}
In video, low image quality always causes weak detection and what Seq-NMS~\cite{han2016seq} does is to raise the confidence of the weak detection according to the high-scoring boxes within same tracklet, which can also be termed as keeping temporal consistency. But imagine such a case where weak detection appears rarely and then Seq-NMS~\cite{han2016seq} will bring no gains. STCA can just make it happen. Specifically, the feature aggregation strategy in STCA is learnable, which means feature can be aggregated adaptively and robustly. After aggregation, proposal's feature will be discriminative enough, especially for low image quality cases, and then the classification gets easier. Hence, it's natural that the Seq-NMS~\cite{han2016seq} in STCA obtains less gain with $T$ getting larger. Specifically, the gain of Seq-NMS~\cite{han2016seq} is 1.4\% and 0.3\% when $T$ is 1 and 31. The extreme case is $T$ equals video length and then it obtains global contextual information in both spatial and temporal dimensions, which has the similar effect to Seq-NMS~\cite{han2016seq}. 
Eventually, in Fig 4, we can easily observe that the box scores are confident and the temporal consistency is well maintained in STCA.

\section{Conclusion and Future Work}

In this work, we propose a novel spatial-temporal context aggregation network for video object detection, which is conceptually simple but effective. In comparison with pixel-wise aggregation methods, our method is proposal-level and naturally robust. Besides, our method can learn the dependency among proposals across spatial-temporal domain, which shows the potential to be end-to-end without extra temporal post-processing.  A more complex design for framework (\eg, adopting more units or the transformer mechanism \cite{dehghani2018universal}) may be investigated in future work.

{\small
\bibliographystyle{unsrt}
\bibliography{egbib}

\begin{thebibliography}{10}

\bibitem{deng2009imagenet}
Jia Deng, Wei Dong, Richard Socher, Li-Jia Li, Kai Li, and Li~Fei-Fei.
\newblock Imagenet: A large-scale hierarchical image database.
\newblock 2009.

\bibitem{redmon2016you}
Joseph Redmon, Santosh Divvala, Ross Girshick, and Ali Farhadi.
\newblock You only look once: Unified, real-time object detection.
\newblock In {\em Proceedings of the IEEE conference on computer vision and
  pattern recognition}, pages 779--788, 2016.

\bibitem{he2017mask}
Kaiming He, Georgia Gkioxari, Piotr Doll{\'a}r, and Ross Girshick.
\newblock Mask r-cnn.
\newblock In {\em Proceedings of the IEEE international conference on computer
  vision}, pages 2961--2969, 2017.

\bibitem{ren2015faster}
Shaoqing Ren, Kaiming He, Ross Girshick, and Jian Sun.
\newblock Faster r-cnn: Towards real-time object detection with region proposal
  networks.
\newblock In {\em Advances in neural information processing systems}, pages
  91--99, 2015.

\bibitem{huang2015densebox}
Lichao Huang, Yi~Yang, Yafeng Deng, and Yinan Yu.
\newblock Densebox: Unifying landmark localization with end to end object
  detection.
\newblock {\em arXiv preprint arXiv:1509.04874}, 2015.

\bibitem{kar2017adascan}
Amlan Kar, Nishant Rai, Karan Sikka, and Gaurav Sharma.
\newblock Adascan: Adaptive scan pooling in deep convolutional neural networks
  for human action recognition in videos.
\newblock In {\em Proceedings of the IEEE conference on computer vision and
  pattern recognition}, pages 3376--3385, 2017.

\bibitem{karpathy2014large}
Andrej Karpathy, George Toderici, Sanketh Shetty, Thomas Leung, Rahul
  Sukthankar, and Li~Fei-Fei.
\newblock Large-scale video classification with convolutional neural networks.
\newblock In {\em Proceedings of the IEEE conference on Computer Vision and
  Pattern Recognition}, pages 1725--1732, 2014.

\bibitem{zhu2017flow}
Xizhou Zhu, Yujie Wang, Jifeng Dai, Lu~Yuan, and Yichen Wei.
\newblock Flow-guided feature aggregation for video object detection.
\newblock In {\em Proceedings of the IEEE International Conference on Computer
  Vision}, pages 408--417, 2017.

\bibitem{bertasius2018object}
Gedas Bertasius, Lorenzo Torresani, and Jianbo Shi.
\newblock Object detection in video with spatiotemporal sampling networks.
\newblock In {\em Proceedings of the European Conference on Computer Vision
  (ECCV)}, pages 331--346, 2018.

\bibitem{wang2018fully}
Shiyao Wang, Yucong Zhou, Junjie Yan, and Zhidong Deng.
\newblock Fully motion-aware network for video object detection.
\newblock In {\em Proceedings of the European Conference on Computer Vision
  (ECCV)}, pages 542--557, 2018.

\bibitem{dosovitskiy2015flownet}
Alexey Dosovitskiy, Philipp Fischer, Eddy Ilg, Philip Hausser, Caner Hazirbas,
  Vladimir Golkov, Patrick Van Der~Smagt, Daniel Cremers, and Thomas Brox.
\newblock Flownet: Learning optical flow with convolutional networks.
\newblock In {\em Proceedings of the IEEE international conference on computer
  vision}, pages 2758--2766, 2015.

\bibitem{dai2017deformable}
Jifeng Dai, Haozhi Qi, Yuwen Xiong, Yi~Li, Guodong Zhang, Han Hu, and Yichen
  Wei.
\newblock Deformable convolutional networks.
\newblock In {\em Proceedings of the IEEE international conference on computer
  vision}, pages 764--773, 2017.

\bibitem{girshick2015fast}
Ross Girshick.
\newblock Fast r-cnn.
\newblock In {\em Proceedings of the IEEE international conference on computer
  vision}, pages 1440--1448, 2015.

\bibitem{vaswani2017attention}
Ashish Vaswani, Noam Shazeer, Niki Parmar, Jakob Uszkoreit, Llion Jones,
  Aidan~N Gomez, {\L}ukasz Kaiser, and Illia Polosukhin.
\newblock Attention is all you need.
\newblock In {\em Advances in Neural Information Processing Systems}, pages
  5998--6008, 2017.

\bibitem{girshick2014rich}
Ross Girshick, Jeff Donahue, Trevor Darrell, and Jitendra Malik.
\newblock Rich feature hierarchies for accurate object detection and semantic
  segmentation.
\newblock In {\em Proceedings of the IEEE conference on computer vision and
  pattern recognition}, pages 580--587, 2014.

\bibitem{dai2016r}
Jifeng Dai, Yi~Li, Kaiming He, and Jian Sun.
\newblock R-fcn: Object detection via region-based fully convolutional
  networks.
\newblock In {\em Advances in neural information processing systems}, pages
  379--387, 2016.

\bibitem{lin2017feature}
Tsung-Yi Lin, Piotr Doll{\'a}r, Ross Girshick, Kaiming He, Bharath Hariharan,
  and Serge Belongie.
\newblock Feature pyramid networks for object detection.
\newblock In {\em Proceedings of the IEEE Conference on Computer Vision and
  Pattern Recognition}, pages 2117--2125, 2017.

\bibitem{liu2016ssd}
Wei Liu, Dragomir Anguelov, Dumitru Erhan, Christian Szegedy, Scott Reed,
  Cheng-Yang Fu, and Alexander~C Berg.
\newblock Ssd: Single shot multibox detector.
\newblock In {\em European conference on computer vision}, pages 21--37.
  Springer, 2016.

\bibitem{lin2017focal}
Tsung-Yi Lin, Priya Goyal, Ross Girshick, Kaiming He, and Piotr Doll{\'a}r.
\newblock Focal loss for dense object detection.
\newblock In {\em Proceedings of the IEEE international conference on computer
  vision}, pages 2980--2988, 2017.

\bibitem{hu2018relation}
Han Hu, Jiayuan Gu, Zheng Zhang, Jifeng Dai, and Yichen Wei.
\newblock Relation networks for object detection.
\newblock In {\em Proceedings of the IEEE Conference on Computer Vision and
  Pattern Recognition}, pages 3588--3597, 2018.

\bibitem{tang2018object}
Peng Tang, Chunyu Wang, Xinggang Wang, Wenyu Liu, Wenjun Zeng, and Jingdong
  Wang.
\newblock Object detection in videos by high quality object linking.
\newblock {\em arXiv preprint arXiv:1801.09823}, 2018.

\bibitem{kang2018t}
Kai Kang, Hongsheng Li, Junjie Yan, Xingyu Zeng, Bin Yang, Tong Xiao, Cong
  Zhang, Zhe Wang, Ruohui Wang, Xiaogang Wang, et~al.
\newblock T-cnn: Tubelets with convolutional neural networks for object
  detection from videos.
\newblock {\em IEEE Transactions on Circuits and Systems for Video Technology},
  28(10):2896--2907, 2018.

\bibitem{han2016seq}
Wei Han, Pooya Khorrami, Tom~Le Paine, Prajit Ramachandran, Mohammad
  Babaeizadeh, Honghui Shi, Jianan Li, Shuicheng Yan, and Thomas~S Huang.
\newblock Seq-{NMS} for video object detection.
\newblock {\em arXiv preprint arXiv:1602.08465}, 2016.

\bibitem{zhu2017deep}
Xizhou Zhu, Yuwen Xiong, Jifeng Dai, Lu~Yuan, and Yichen Wei.
\newblock Deep feature flow for video recognition.
\newblock In {\em Proceedings of the IEEE Conference on Computer Vision and
  Pattern Recognition}, pages 2349--2358, 2017.

\bibitem{wang2018non}
Xiaolong Wang, Ross Girshick, Abhinav Gupta, and Kaiming He.
\newblock Non-local neural networks.
\newblock In {\em Proceedings of the IEEE Conference on Computer Vision and
  Pattern Recognition}, pages 7794--7803, 2018.

\bibitem{he2016deep}
Kaiming He, Xiangyu Zhang, Shaoqing Ren, and Jian Sun.
\newblock Deep residual learning for image recognition.
\newblock In {\em Proceedings of the IEEE conference on computer vision and
  pattern recognition}, pages 770--778, 2016.

\bibitem{shrivastava2016training}
Abhinav Shrivastava, Abhinav Gupta, and Ross Girshick.
\newblock Training region-based object detectors with online hard example
  mining.
\newblock In {\em Proceedings of the IEEE Conference on Computer Vision and
  Pattern Recognition}, pages 761--769, 2016.

\bibitem{chen2015mxnet}
Tianqi Chen, Mu~Li, Yutian Li, Min Lin, Naiyan Wang, Minjie Wang, Tianjun Xiao,
  Bing Xu, Chiyuan Zhang, and Zheng Zhang.
\newblock Mxnet: A flexible and efficient machine learning library for
  heterogeneous distributed systems.
\newblock {\em arXiv preprint arXiv:1512.01274}, 2015.

\bibitem{xiao2018video}
Fanyi Xiao and Yong Jae~Lee.
\newblock Video object detection with an aligned spatial-temporal memory.
\newblock In {\em Proceedings of the European Conference on Computer Vision
  (ECCV)}, pages 485--501, 2018.

\bibitem{feichtenhofer2017detect}
Christoph Feichtenhofer, Axel Pinz, and Andrew Zisserman.
\newblock Detect to track and track to detect.
\newblock In {\em Proceedings of the IEEE International Conference on Computer
  Vision}, pages 3038--3046, 2017.

\bibitem{chen2018optimizing}
Kai Chen, Jiaqi Wang, Shuo Yang, Xingcheng Zhang, Yuanjun Xiong, Chen
  Change~Loy, and Dahua Lin.
\newblock Optimizing video object detection via a scale-time lattice.
\newblock In {\em Proceedings of the IEEE Conference on Computer Vision and
  Pattern Recognition}, pages 7814--7823, 2018.

\bibitem{dehghani2018universal}
Mostafa Dehghani, Stephan Gouws, Oriol Vinyals, Jakob Uszkoreit, and {\L}ukasz
  Kaiser.
\newblock Universal transformers.
\newblock In {\em ICLR}, 2019.

\end{thebibliography}
}

\end{document}